\documentclass[runningheads]{llncs}

\usepackage[T1]{fontenc}
\usepackage{graphicx}
\usepackage{amsmath}
\usepackage{array}
\usepackage{booktabs}
\usepackage[table]{xcolor}
\usepackage{microtype}
\usepackage{url}
\definecolor{oursrow}{RGB}{255,235,175}
\definecolor{baselinerow}{RGB}{226,239,218}
\definecolor{mainbaselinerow}{RGB}{220,235,250}
\definecolor{ablationrow}{RGB}{220,235,250}
\definecolor{testrow}{RGB}{255,235,175}
\newcommand{\tick}{\raisebox{0.15ex}{\tiny$\surd$}}

\begin{document}
\pagestyle{empty}

\title{CIRSeg: Coarse-to-Fine Intensity-Robust Liver Segmentation with Source-Free Continual Test-Time Adaptation}
\titlerunning{CIRSeg for CARE 2026 Liver Segmentation}

\author{Ruoshi Xu\inst{2} \and
        Mingqi Gao\inst{3} \and
        Shengda Luo\inst{2}\textsuperscript{*} \and
        Jingkun Chen\inst{1}\textsuperscript{*}}
\authorrunning{Xu et al.}
\institute{School of Automation, Northwestern Polytechnical University, Xi'an, China\inst{1}\\
           \email{jingkunchen@nwpu.edu.cn}\\
           Chinese Medicine Guangdong Laboratory, Zhuhai, China\inst{2}\\
           \email{xuruoshi2003@gmail.com, luoshengda@hqcmlab.cn}\\
           School of Computer Science, University of Sheffield, Sheffield, UK\inst{3}\\
           \email{m.gao@sheffield.ac.uk}}

\maketitle
\begingroup
\renewcommand{\thefootnote}{*}
\footnotetext{Corresponding author.}
\endgroup

\begin{abstract}
Reliable liver segmentation in contrast-enhanced MRI is essential for quantitative hepatic assessment, treatment planning, and longitudinal disease monitoring. However, limited annotated data and scanner- or vendor-dependent intensity variations can cause overfitting and poor generalization to unseen acquisition domains. Moreover, simultaneously achieving robust global localization and precise boundary delineation remains challenging, while predictions may contain isolated false-positive regions outside the main liver component. To address these challenges, we propose CIRSeg, a coarse-to-fine, intensity-robust liver segmentation framework based on nnU-Netv2. CIRSeg combines 3D CutMix with stochastic intensity transfer using either Nyul augmentation or histogram matching to improve robustness to heterogeneous MRI intensities. Its cascaded architecture decouples low-resolution anatomical localization from full-resolution boundary refinement. At inference, source-free test-time adaptation based on confidence-filtered predictions and probability-prior regularization further improves robustness to out-of-distribution inputs. As a final deterministic post-processing step, largest connected component filtering removes isolated false-positive regions. On the CARE 2026 test set, CIRSeg achieves Dice scores of 97.13\% and 97.93\% on the in-domain and unseen-domain subsets, with corresponding HD95 values of 20.18 mm and 11.30 mm, respectively. These results demonstrate consistently accurate segmentation across both in-domain and unseen acquisition settings. The code is available at \url{https://github.com/jingkunchen/MICCAI_CARE_2026}

\end{abstract}

\section{Introduction}

Contrast-enhanced magnetic resonance imaging (MRI) is widely used for quantitative liver assessment, treatment planning, surgical navigation, and longitudinal disease monitoring. However, manual delineation of the liver in three-dimensional MRI volumes is labor-intensive and requires substantial expert knowledge. Reliable automated segmentation remains challenging because annotated training volumes are limited, while abdominal MRI intensities vary considerably across scanner vendors, acquisition protocols, reconstruction settings, contrast enhancement patterns, and patients. Such variations can cause models to overfit to cohort-specific appearances and compromise generalization to unseen acquisition domains. In addition, liver segmentation requires both robust global localization and precise boundary delineation, which rely on different spatial contexts, while predicted masks may contain small disconnected false-positive regions outside the dominant liver structure.

Recent medical image segmentation studies have addressed these challenges from complementary perspectives. Report-supervised, weakly-supervised, and pseudo-labeling methods reduce dependence on dense annotations~\cite{bassi2025learning,hu2025better,chen2025gaze,chen2025addressing}, while foundation-model studies investigate the transferability of general-purpose segmentation priors to medical images~\cite{gu2025sam}. Other approaches focus on domain generalization~\cite{cheng2025dynamic} and test-time adaptation~\cite{chen2025grata,ji2026tact} to improve robustness when deployment data differ from the labeled training distribution. Nevertheless, these directions are often investigated independently, whereas robust liver segmentation under heterogeneous acquisition conditions requires coordinated treatment of limited training diversity, intensity-domain variation, multi-scale anatomical modeling, and inference-time distribution shift.

To address these challenges, we propose CIRSeg, a coarse-to-fine, intensity-robust liver segmentation framework built upon nnU-Netv2~\cite{isensee2021nnunet}. At the data level, CIRSeg introduces 3D CutMix~\cite{yun2019cutmix} to improve regularization and increase spatial diversity under limited labeled data. We further design a composite intensity-domain augmentation strategy that probabilistically applies either Nyul augmentation~\cite{nyul2000standardization} or histogram matching~\cite{gonzalez2018dip}. For histogram matching, the reference volume is randomly sampled from the training set, allowing the model to encounter a broader range of plausible intensity distributions during training. Together, these augmentations reduce sensitivity to acquisition-specific appearance patterns and improve generalization to heterogeneous domains.

At the model level, CIRSeg adopts cascaded nnU-Netv2. A low-resolution network first performs coarse liver localization, and a subsequent full-resolution network refines local structures and boundaries. This coarse-to-fine formulation separates global anatomical localization from fine-grained segmentation refinement. During full-resolution inference, CIRSeg incorporates a source-free continual test-time adaptation strategy inspired by SicTTA~\cite{wu2026sictta}. The model is continually updated across test cases using confidence-filtered anchor predictions and probability-prior regularization, improving robustness to out-of-distribution inputs while limiting adaptation drift. Finally, a deterministic three-dimensional largest connected component post-processing step removes isolated false-positive regions and retains the dominant liver component.

Our contributions are threefold: \textbf{1.} We develop an intensity-robust training data augmentation recipe that combines 3D CutMix with probabilistic Nyul augmentation and histogram matching to increase spatial and intensity diversity and improve generalization under limited annotations. \textbf{2.} We construct a coarse-to-fine liver segmentation framework based on cascaded nnU-Netv2 and integrate it with source-free continual test-time adaptation to improve cross-domain robustness while limiting adaptation drift. \textbf{3.} CIRSeg achieves strong performance on the CARE 2026 test set, demonstrating accurate liver segmentation and robust generalization to unseen acquisition domains.

\section{Method}

\begin{figure}
\centering
\includegraphics[width=\textwidth]{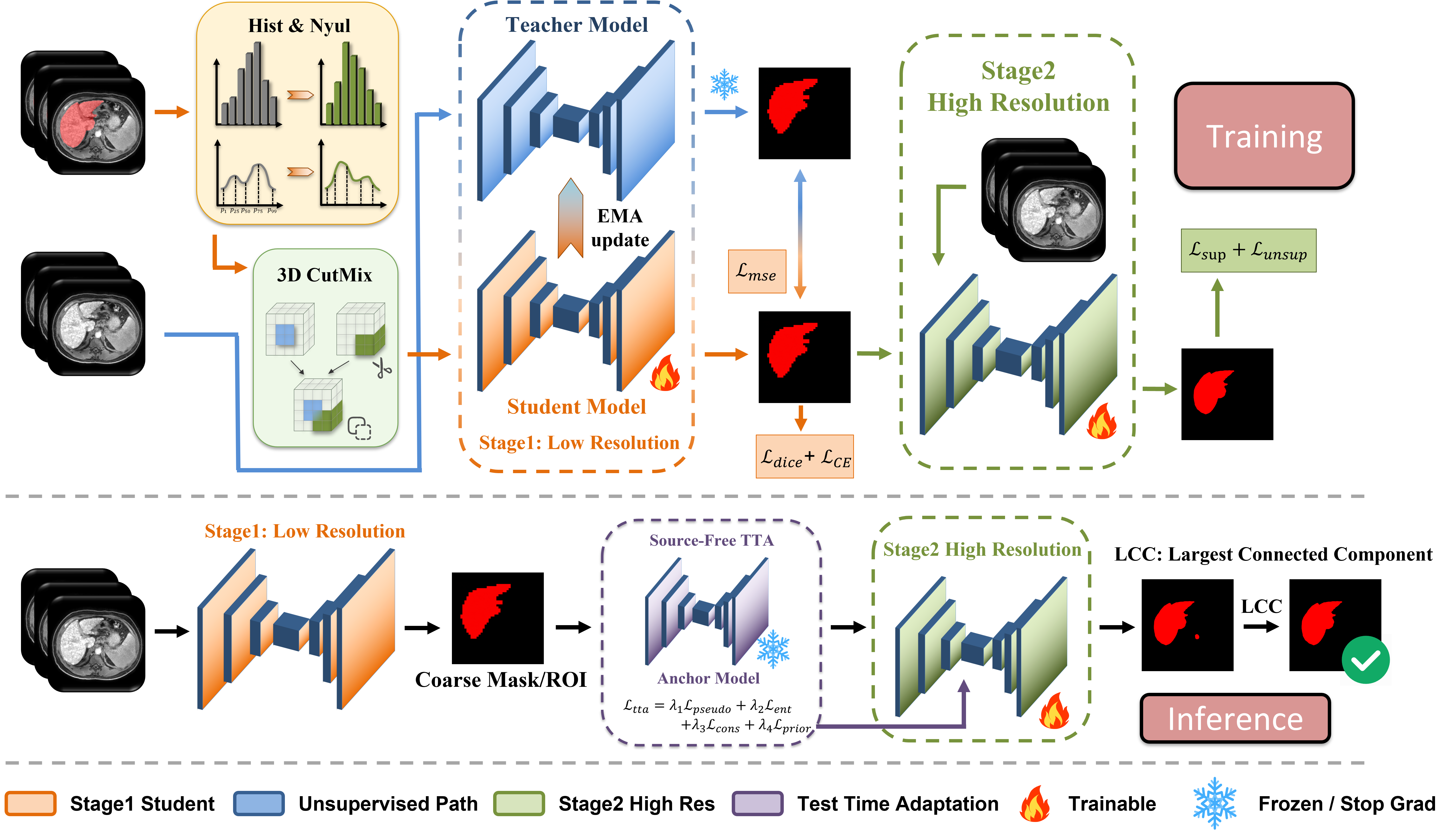}
\caption{Overall architecture of CIRSeg.}
\label{fig:pipeline}
\end{figure}

Figure~\ref{fig:pipeline} shows the overall architecture of CIRSeg.

\subsection{Baseline}

We build our baseline upon CoSSeg-TTA~\cite{lou2026cosseg}, the first-place
solution of the CARE Liver 2025 challenge. It employs nnU-Netv2
~\cite{isensee2021nnunet} as the segmentation backbone and incorporates
mean-teacher semi-supervised learning~\cite{tarvainen2017mean} and histogram
matching transfer. We retain these components and focus on improving data
diversity, coarse-to-fine segmentation, and inference-time robustness.

\subsection{Composite Data Augmentation}

The CARE 2026 training set contains only a limited number of annotated volumes,
while MRI appearances vary substantially across scanners, vendors, and
acquisition protocols. We therefore combine 3D CutMix with probabilistic
intensity-domain augmentation. Unlabeled training volumes are used for
semi-supervised consistency learning and may also serve as reference volumes
for intensity transfer.

\paragraph{3D CutMix.}
To improve regularization under limited annotations, we apply 3D CutMix during
supervised training. Given two labeled samples $(x_a,y_a)$ and $(x_b,y_b)$, a
3D cuboid $B$ is randomly sampled. Let $M_B$ be a binary mask whose value is
one inside $B$. The mixed image and label are defined as
\begin{equation}
x_{\mathrm{mix}}
=
(1-M_B)\odot x_a + M_B\odot x_b,
\end{equation}
\begin{equation}
y_{\mathrm{mix}}
=
(1-M_B)\odot y_a + M_B\odot y_b,
\end{equation}
where $\odot$ denotes element-wise multiplication. Applying the same cuboid to
the image and label preserves voxel-wise supervision while increasing local
appearance and boundary diversity.

\paragraph{Histogram matching and Nyul augmentation.}
For intensity-domain robustness, we combine histogram matching with Nyul
augmentation. For each selected source volume, either transformation is applied
probabilistically, and the reference volume is randomly sampled from the entire
CARE 2026 training set, including both labeled and unlabeled cases. Both
transformations modify only the image intensities and leave the segmentation
label unchanged.

\begin{figure}[!ht]
\centering
\includegraphics[width=\textwidth]{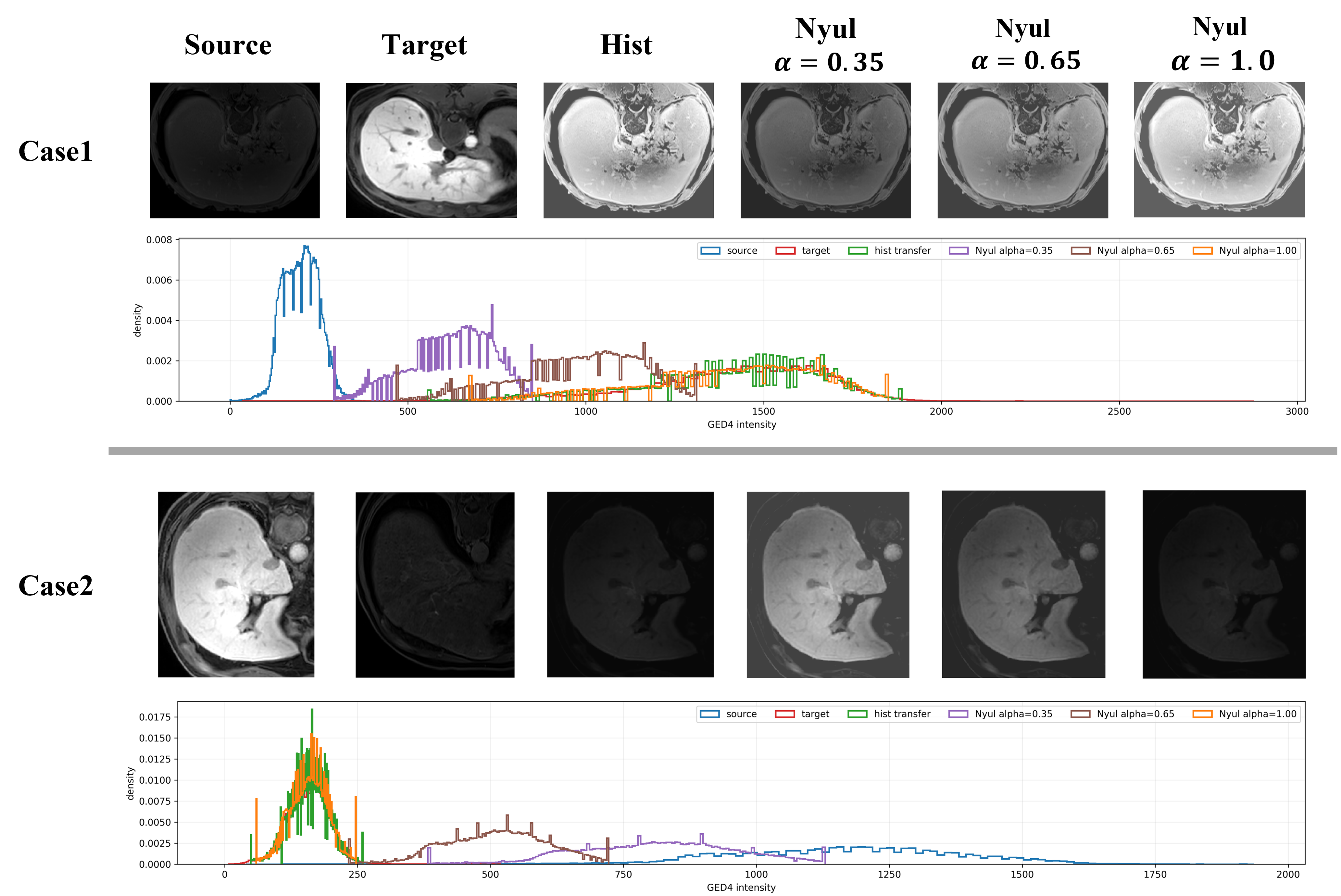}
\caption{Visualization of histogram matching and Nyul augmentation. Each row
shows a source volume, a randomly sampled reference volume, the
histogram-matched result, and Nyul-augmented results with different transfer
strengths $\alpha$, together with their foreground intensity distributions.}
\label{fig:intensity_aug}
\end{figure}

A body foreground mask is first obtained by intensity thresholding, and all
intensity statistics are computed within this region. Given the foreground
voxels of a source volume $S$ and a reference volume $R$, their percentile
landmarks are defined as
\begin{equation}
S_Q=[\mathcal{Q}_{q_1}(S),\mathcal{Q}_{q_2}(S),\ldots,
\mathcal{Q}_{q_K}(S)],
\end{equation}
\begin{equation}
R_Q=[\mathcal{Q}_{q_1}(R),\mathcal{Q}_{q_2}(R),\ldots,
\mathcal{Q}_{q_K}(R)],
\end{equation}
where $\mathcal{Q}_{q_k}(\cdot)$ denotes the $q_k$-th foreground intensity
percentile. The Nyul mapping function $g(\cdot)$ is defined by piecewise-linear
interpolation between corresponding landmarks. For
$z\in[S_Q[k],S_Q[k+1]]$,
\begin{equation}
g(z)=R_Q[k]+
\frac{z-S_Q[k]}{S_Q[k+1]-S_Q[k]}
\left(R_Q[k+1]-R_Q[k]\right).
\end{equation}
Values outside the landmark range are clipped to the nearest boundary landmark.
The mapped volume is blended with the original source volume:
\begin{equation}
S_{\mathrm{aug}}=(1-\alpha)S+\alpha g(S),
\end{equation}
where $\alpha$ controls the transfer strength. Histogram matching directly
transfers the source intensity distribution toward that of the sampled
reference, whereas Nyul augmentation generates intermediate appearances by
varying $\alpha$. This composite strategy broadens the sampled intensity-domain
space and improves robustness to heterogeneous acquisitions.
Figure~\ref{fig:intensity_aug} visualizes the resulting intensity variations.

\subsection{Cascaded Training}

Global liver localization and fine-grained boundary delineation require
different spatial contexts. We therefore adopt the standard cascaded
nnU-Netv2 training strategy. A low-resolution model first predicts the coarse
liver extent:
\begin{equation}
p_{\mathrm{low}}=f_{\mathrm{low}}(x_{\mathrm{down}}),
\end{equation}
where $x_{\mathrm{down}}$ denotes the resampled low-resolution volume. The
prediction is then resampled to the original image space and provided to the
full-resolution cascade:
\begin{equation}
p_{\mathrm{cas}}
=
f_{\mathrm{cas}}([x,\operatorname{up}(p_{\mathrm{low}})]),
\end{equation}
where $[\cdot,\cdot]$ denotes channel-wise concatenation. The low-resolution
prediction provides a coarse anatomical prior, allowing the full-resolution
model to focus on local structures and liver boundaries.

\subsection{Inference-Time Adaptation and Refinement}

\paragraph{Source-free continual test-time adaptation.}
During inference, the low-resolution model first produces a coarse liver
prediction without adaptation. Source-free continual test-time adaptation,
inspired by SicTTA~\cite{wu2026sictta}, is then applied to the full-resolution
cascade. A frozen anchor $f_a$ and an adaptive student $f_s$ are initialized
from the trained cascade model. For each test batch, the anchor produces
$p_a=\mathrm{softmax}(f_a(x))$. This is a binary segmentation task with
$C=2$ classes, background and liver foreground; $p_a^k(v)$ denotes the
probability of class $k\in\{1,2\}$ at voxel $v$. The anchor pseudo-label is
$\hat y_a(v)=\arg\max_k p_a^k(v)$, and its voxel-wise confidence is
$c(v)=p_a^{\hat y_a(v)}(v)=\max_{k\in\{1,2\}}p_a^k(v)$.

To filter unreliable updates, we compute a confidence-consistency (CC) score
from at most 2048 randomly sampled anchor probability vectors. Specifically,
let $\Omega$ denote the set of all voxels in the current test batch and let
$v_1,\ldots,v_N\in\Omega$ be the sampled voxels, with
$N\le\min(2048,|\Omega|)$. Then $P\in{R}^{N\times C}$
stacks $p_a(v_n)$ as rows and $\tilde P$ is obtained by row-wise $\ell_2$
normalization. Their class-correlation matrix is then
$R=\tilde P^\top\tilde P$, and
\begin{equation}
s_{\mathrm{CC}}
=-\frac{1}{C}\sum_{i,j}\pi_{ij}\log(\pi_{ij}+\epsilon),
\qquad
\pi=\mathrm{softmax}(R),
\end{equation}
where softmax is applied row-wise, so each row of $\pi$ is a distribution over
classes. Thus, $s_{\mathrm{CC}}$ is the mean entropy of the row-wise
class-correlation distributions; a lower value indicates a more concentrated
and consistent correlation pattern. We maintain the most recent 32 CC scores.
For the first four batches, reliability requires mean confidence
$\bar c\ge0.85$; afterwards, a batch is reliable if either its CC score falls
below the low-tail cutoff of the history or $\bar c\ge0.85$. Only batches passing the criteria are used for adaptation.

The prior bank is updated only for batches satisfying both reliability and
minimum high-confidence-mask-ratio criteria, by appending the mean
high-confidence anchor probability and retaining the most recent 32 entries
in FIFO order. Let $\bar q$ denote
the mean bank distribution and
$q_s=|\Omega|^{-1}\sum_{v\in\Omega}p_s(v)$ the mean student distribution.
We define
\begin{equation}
\mathcal{L}_{\mathrm{prior}}=\|q_s-\bar q\|_2^2,
\end{equation}
which constrains global class proportions rather than providing voxel-wise
supervision. The student is optimized with
\begin{equation}
\mathcal{L}_{\mathrm{tta}}
=
\lambda_{\mathrm{pseudo}}\mathcal{L}_{\mathrm{pseudo}}
+\lambda_{\mathrm{ent}}\mathcal{L}_{\mathrm{ent}}
+\lambda_{\mathrm{cons}}\mathcal{L}_{\mathrm{cons}}
+\lambda_{\mathrm{prior}}\mathcal{L}_{\mathrm{prior}},
\end{equation}
where the first three terms denote high-confidence pseudo-label cross-entropy,
entropy minimization, and anchor--student KL consistency, respectively.
The student is not reset between test volumes; its adapted parameters are
carried forward throughout the test stream, while the anchor remains frozen.

\paragraph{3D largest connected component filtering.}
Post-processing is implemented as standard, deterministic three-dimensional
largest connected component filtering. The adapted cascade prediction is
converted into a binary liver mask. Given its connected components
$\{C_k\}_{k=1}^{K}$, the final prediction is
\begin{equation}
M_{\mathrm{pp}}=C_{\arg\max_k |C_k|}.
\end{equation}
If the prediction contains at most one foreground component, it is left
unchanged. This operation removes isolated false-positive regions and improves
anatomical plausibility, particularly for surface-distance metrics such as
HD95.

\section{Experiments}

\subsection{Dataset and Evaluation Metrics}

We participate in the GED4 Contrast-Enhanced LiSeg Subtask of the CARE 2026
Liver Track and use the official dataset released for this subtask
~\cite{care2026,liu2026lifs,liu2025merit,gao2023reliable,wu2023meru}.
Following the official challenge protocol and dataset splits, we use the
provided training, validation, and test sets for model development and
evaluation. No additional public or private datasets are used.

The official evaluation metrics are Dice score and the 95th-percentile
Hausdorff distance (HD95). Higher Dice and lower HD95 indicate better
segmentation performance.

\subsection{Implementation Setup}

All experiments are conducted on one NVIDIA RTX 6000 24GB GPU. Both Stages of nnU-Netv2
(stage1 low-resolution and Stage2 high-resolution) are optimized using SGD
with the same mean-teacher objective
\begin{equation}
\mathcal{L}_{\mathrm{train}}^{(s)}
=
\mathcal{L}_{\mathrm{sup}}^{(s)}
+
\lambda_{\mathrm{u}}(t)\mathcal{L}_{\mathrm{unsup}}^{(s)},
\quad s\in\{\mathrm{low},\mathrm{cas}\},
\end{equation}
where $\lambda_{\mathrm{u}}(t)$ controls the consistency weight. The two stages
differ in input resolution and cascade conditioning but use identical loss
definitions. The supervised term is the combined Dice and cross-entropy loss
\begin{equation}
\mathcal{L}_{\mathrm{sup}}
=\mathcal{L}_{\mathrm{Dice}}+\mathcal{L}_{\mathrm{CE}},
\end{equation}
where
\begin{equation}
\mathcal{L}_{\mathrm{Dice}}
=1-\frac{2\sum_{i=1}^{N}p_i y_i+\epsilon}
{\sum_{i=1}^{N}p_i+\sum_{i=1}^{N}y_i+\epsilon},
\end{equation}
\begin{equation}
\mathcal{L}_{\mathrm{CE}}
=-\frac{1}{N}\sum_{i=1}^{N}
\left[y_i\log p_i+(1-y_i)\log(1-p_i)\right].
\end{equation}
The unsupervised term is the mean-squared consistency loss between student and
EMA-teacher predictions on unlabeled data:
\begin{equation}
\mathcal{L}_{\mathrm{unsup}}^{(s)}
=
\frac{1}{N}\sum_{i=1}^{N}
\left\|p_{i,s}^{\mathrm{stu}}-p_{i,s}^{\mathrm{tea}}\right\|_2^2.
\end{equation}
Here, $p_i$ denotes the predicted foreground probability at voxel $i$, $y_i$
is the corresponding binary ground-truth label, $N$ is the number of voxels,
and $\epsilon$ is a smoothing constant. The initial learning rate is $10^{-2}$,
the weight decay is $3\times10^{-5}$, and training runs for 150 epochs with 250
iterations per epoch. The consistency term is activated after epoch 40 with an
80-epoch sigmoid ramp-up, and the teacher EMA decay is set to 0.95.

For augmentation, 3D CutMix is applied with probability 0.5. The mixing ratio
is sampled from $\mathrm{Beta}(1.0,1.0)$, and the CutMix box side ratio is
clipped to $[0.20,0.55]$. Histogram matching and Nyul augmentation are selected
with probabilities 0.7 and 0.3, respectively. The Nyul branch uses 17
foreground quantile landmarks from the 0.5th to 99.5th percentiles and samples
the blending coefficient from $\alpha\in[0.2,0.8]$.

The cascade follows the \texttt{nnU-Netv2 3d\_lowres} and
\texttt{3d\_cascade\_fullres} configurations. During inference, source-free
continual test-time adaptation is disabled for the low-resolution stage and
enabled for the full-resolution cascade stage. The TTA optimizer uses a
learning rate of $10^{-5}$, and the loss weights are set as
\begin{equation}
\mathcal{L}_{tta}
=
1.0\mathcal{L}_{pseudo}
+
0.1\mathcal{L}_{ent}
+
0.2\mathcal{L}_{cons}
+
0.05\mathcal{L}_{prior}.
\end{equation}
The confidence threshold is 0.85, the consistency confidence threshold is
0.90, the minimum foreground fraction is 0.0005, and the memory bank size is
32. Largest connected component filtering uses 3D connectivity.

\subsection{Main Experiment}
\label{sec:main_experiment}

Table~\ref{tab:baseline} compares CIRSeg with representative 3D medical image
segmentation methods using five-fold cross-validation on the 30 annotated
training cases. For a fair comparison, all methods are trained exclusively on
the official training set without pretrained models. CoSSeg-TTA and CIRSeg use both labeled and unlabeled
cases through their semi-supervised learning components, whereas the other
methods are trained only on labeled cases because their original designs do
not provide a mechanism for learning from unlabeled data. CIRSeg achieves the best overall performance among the
evaluated methods, obtaining the highest Dice score and the lowest HD95. The
reproduced CoSSeg-TTA~\cite{lou2026cosseg} framework is used as the baseline
for further ablation analysis.

\begin{table}[t]
\centering
\caption{Main experimental comparison using five-fold cross-validation on
the official training set.}
\label{tab:baseline}
\begin{tabular}{lcc}
\toprule
Method & Dice $\uparrow$ & HD95 $\downarrow$ \\
\midrule
\rowcolor{oursrow}
\textbf{CIRSeg (ours)} & \textbf{0.9595} & \textbf{5.29} \\
\rowcolor{mainbaselinerow}
CoSSeg-TTA Baseline & 0.9445 & 7.74 \\
\rowcolor{mainbaselinerow}
nnU-Netv2 & 0.8973 & 26.95 \\
\rowcolor{mainbaselinerow}
Attention U-Net & 0.8713 & 46.21 \\
\rowcolor{mainbaselinerow}
SegResNet & 0.8704 & 32.19 \\
\rowcolor{mainbaselinerow}
3D U-Net & 0.8537 & 37.07 \\
\rowcolor{mainbaselinerow}
Swin UNETR & 0.8499 & 34.48 \\
\rowcolor{mainbaselinerow}
V-Net & 0.8484 & 28.26 \\
\bottomrule
\end{tabular}
\end{table}

\paragraph{Qualitative Analysis.}

\begin{figure}[t]
\centering
\includegraphics[width=\textwidth]{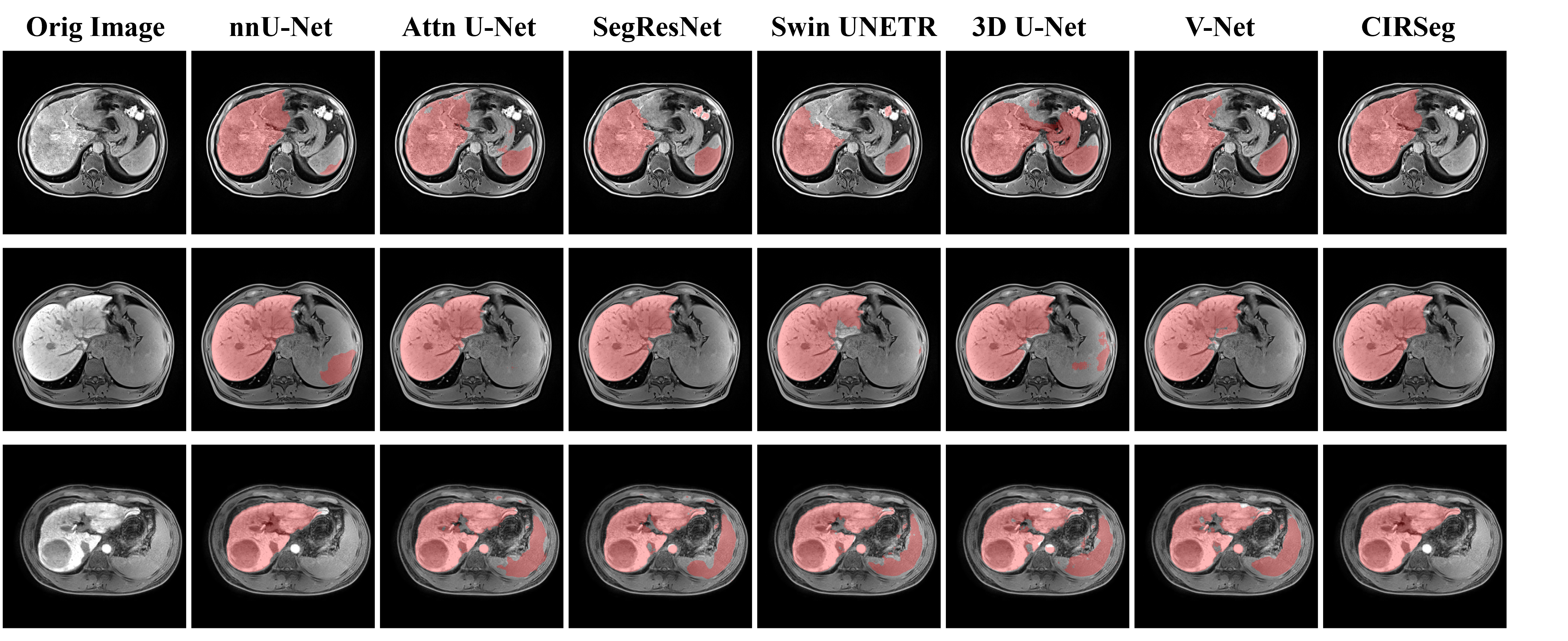}
\caption{Qualitative comparison on representative CARE 2026 validation cases.
The last column shows CIRSeg predictions, which provide more complete liver
coverage and no disconnected false-positive regions compared with previous
methods.}
\label{fig:quali}
\end{figure}

Figure~\ref{fig:quali} presents qualitative comparisons on representative
cases from the CARE 2026 validation set. Since the validation annotations are
not publicly available, we focus on visual comparison of the predicted liver
masks. CIRSeg produces more complete and anatomically consistent liver regions
compared with previous approaches. In particular, the final predictions
preserve smoother boundaries, better capture irregular liver contours, and
reduce disconnected false-positive regions.

\subsection{Ablation Study}

Table~\ref{tab:ablation} presents the cumulative ablation results obtained by
submitting validation and test predictions to the official CARE 2026 evaluation
website. The baseline already achieves strong Dice performance, while the relatively
high HD95 indicates that boundary errors and isolated false-positive regions
remain important failure modes.

Adding largest connected component filtering reduces HD95 from 24.69 to 20.61,
demonstrating that anatomical connectivity constraints effectively remove
distant false-positive components. Introducing 3D CutMix further improves
boundary robustness and reduces HD95 to 19.18. The composite intensity
augmentation strategy, combining histogram matching and Nyul augmentation,
achieves additional improvement, indicating that expanding the sampled
intensity-domain space improves robustness against heterogeneous MRI
appearances.

The cascaded nnU-Netv2 architecture further reduces HD95 to 18.31 by providing
a coarse anatomical localization prior for full-resolution refinement. The
source-free continual test-time adaptation module does not improve validation
performance, whose cases are mainly in-domain. TTA-enabled predictions are
lower on ID Test1/2 but higher on OOD, suggesting domain-dependent utility.
Because all test submissions used TTA, no paired no-TTA predictions were
available for causal comparison. The official test-set evaluation, including
case ordering, was performed by the organizers after model submission and was
outside our control; therefore, the ordering sensitivity of continual TTA
could not be evaluated.

\begin{table}[t]
\centering
\caption{Cumulative ablation results obtained through the official CARE 2026
evaluation website. The \textbf{baseline} is the reproduced
\textbf{CoSSeg-TTA} framework described in
Section~\ref{sec:main_experiment}. \textbf{LCC}: largest connected component
filtering; \textbf{Composite Aug}: composite intensity augmentation using
histogram matching and Nyul augmentation; \textbf{Cascade}: cascaded
\textbf{nnU-Netv2} architecture; \textbf{TTA}: source-free continual test-time
adaptation.}
\label{tab:ablation}
\scriptsize
\resizebox{\textwidth}{!}{
\begin{tabular}{|l|c|c|c|c|c|c|c|}
\hline
Exp & LCC & CutMix & Composite Aug & Cascade & TTA &
Dice $\uparrow$ & HD95 $\downarrow$\\
\hline
\rowcolor{baselinerow}
Baseline &  &  &  &  &  & 0.9693 & 24.69\\
\rowcolor{ablationrow}
A & \tick &  &  &  &  & 0.9708 & 20.61\\
\rowcolor{ablationrow}
B & \tick & \tick &  &  &  & 0.9702 & 19.18\\
\rowcolor{ablationrow}
C & \tick & \tick & \tick &  &  & 0.9715 & 18.93\\
\rowcolor{ablationrow}
D & \tick & \tick & \tick & \tick &  & 0.9713 & 18.31\\
\rowcolor{ablationrow}
E & \tick & \tick & \tick & \tick & \tick & 0.9713 & 18.31\\
\hline
\rowcolor{testrow}
ID Test1 & \tick & \tick & \tick & \tick &  & \textbf{0.9713} & \textbf{20.18}\\
\rowcolor{testrow}
ID Test2 & \tick & \tick & \tick & \tick & \tick & \textbf{0.9709} & \textbf{20.85}\\
\rowcolor{testrow}
OOD Test & \tick & \tick & \tick & \tick & \tick & \textbf{0.9797} & \textbf{11.36}\\
\hline
\end{tabular}}
\end{table}

\section{Conclusion}

In this work, we present CIRSeg, an intensity-robust coarse-to-fine liver
segmentation framework for the CARE 2026 Liver Segmentation Track. CIRSeg
addresses the challenges of limited annotated data and heterogeneous MRI
appearance variations through a composite augmentation strategy combining
3D CutMix, histogram matching, and Nyul augmentation. It further integrates a
cascaded nnU-Netv2 architecture for separating coarse anatomical localization
from fine boundary refinement, together with source-free continual test-time
adaptation to improve robustness under distribution shifts. A deterministic
three-dimensional largest connected component filter is applied as the final
post-processing step to remove isolated false positives. Extensive experiments
on the CARE 2026 dataset demonstrate that CIRSeg achieves accurate segmentation
performance on both in-domain and unseen-domain test subsets, validating the
effectiveness of the proposed training and inference strategies for robust liver
segmentation across heterogeneous MRI acquisitions.

\section*{Acknowledgements}
This work was supported by the National Natural Science Foundation of China
(Grant No. 82505358), Innovation Team and Talents Cultivation Program of
National Administration of Traditional Chinese Medicine. (No:
ZYYCXTD-D-202403), and the Scientiffc Research Start-up Funds of the Chinese
Medicine Guangdong Laboratory (Grant No. HQL2025SU011).

\bibliographystyle{splncs04}
\bibliography{main}

@article{isensee2021nnunet,
  author  = {Isensee, Fabian and Jaeger, Paul F. and Kohl, Simon A. A. and Petersen, Jens and Maier-Hein, Klaus H.},
  title   = {{nnU-Net}: A Self-Configuring Method for Deep Learning-Based Biomedical Image Segmentation},
  journal = {Nature Methods},
  year    = {2021},
  volume  = {18},
  number  = {2},
  pages   = {203--211},
  doi     = {10.1038/s41592-020-01008-z}
}

@inproceedings{yun2019cutmix,
  author    = {Yun, Sangdoo and Han, Dongyoon and Oh, Seong Joon and Chun, Sanghyuk and Choe, Junsuk and Yoo, Youngjoon},
  title     = {{CutMix}: Regularization Strategy to Train Strong Classifiers with Localizable Features},
  booktitle = {Proceedings of the IEEE/CVF International Conference on Computer Vision},
  pages     = {6023--6032},
  year      = {2019},
  doi       = {10.1109/ICCV.2019.00612},
  eprint    = {1905.04899},
  archivePrefix = {arXiv},
  primaryClass = {cs.CV}
}

@article{nyul2000standardization,
  author  = {Nyul, Laszlo G. and Udupa, Jayaram K. and Zhang, Xuan},
  title   = {New Variants of a Method of {MRI} Scale Standardization},
  journal = {IEEE Transactions on Medical Imaging},
  year    = {2000},
  volume  = {19},
  number  = {2},
  pages   = {143--150},
  doi     = {10.1109/42.836373}
}

@book{gonzalez2018dip,
  author    = {Gonzalez, Rafael C. and Woods, Richard E.},
  title     = {Digital Image Processing},
  edition   = {4},
  publisher = {Pearson},
  address   = {New York},
  year      = {2018},
  isbn      = {9780133356724}
}

@inproceedings{tarvainen2017mean,
  author    = {Tarvainen, Antti and Valpola, Harri},
  title     = {Mean Teachers Are Better Role Models: Weight-Averaged Consistency Targets Improve Semi-Supervised Deep Learning Results},
  booktitle = {Advances in Neural Information Processing Systems},
  volume    = {30},
  year      = {2017}
}

@article{wu2026sictta,
  author  = {Wu, Jianghao and Liu, Xinya and Wang, Guotai and Zhang, Shaoting},
  title   = {{SicTTA}: Single Image Continual Test Time Adaptation for Medical Image Segmentation},
  journal = {Medical Image Analysis},
  year    = {2026},
  volume  = {108},
  pages   = {103859},
  doi     = {10.1016/j.media.2025.103859}
}

@misc{care2026,
  author    = {{CARE 2026 Organizers}},
  title     = {{CARE} 2026: Comprehensive Analysis of {RE}al-World Medical Images -- Heart and Liver},
  howpublished = {\url{https://www.zmic.org.cn/care_2026/}},
  year      = {2026},
  note      = {Accessed: 2026-07-21}
}

@misc{liu2026lifs,
  author        = {Liu, Yuanye and Shi, Nannan and Zhang, Zhejia and Zhang, Hanxiao and Wang, Boya and Yu, Derong and Wang, Nao and Jin, Yuxin and Zhou, Yang and Yuan, Kunhao and Wang, Siqi and Yang, Lida and Qiao, Xu and Liu, Wentao and He, Xuelei and Hong, Xin and Zheng, Guoyan and Chen, Xin and Yang, Guang-Zhong and Zhang, Le and Li, Lei and Shi, Yuxin and Zhuang, Xiahai},
  title         = {How Far Has {AI} Come in Liver Fibrosis Staging? A Large-Scale Real-World Dataset and Benchmark},
  year          = {2026},
  eprint        = {2605.25595},
  archivePrefix = {arXiv},
  primaryClass  = {cs.CV},
  url           = {https://arxiv.org/abs/2605.25595}
}

@article{liu2025merit,
  author  = {Liu, Yuanye and Gao, Zheyao and Shi, Nannan and Wu, Fuping and Shi, Yuxin and Chen, Qingchao and Zhuang, Xiahai},
  title   = {{MERIT}: Multi-View Evidential Learning for Reliable and Interpretable Liver Fibrosis Staging},
  journal = {Medical Image Analysis},
  volume  = {102},
  pages   = {103507},
  year    = {2025}
}

@inproceedings{gao2023reliable,
  author    = {Gao, Zheyao and Liu, Yuanye and Wu, Fuping and Shi, Nannan and Shi, Yuxin and Zhuang, Xiahai},
  title     = {A Reliable and Interpretable Framework of Multi-View Learning for Liver Fibrosis Staging},
  booktitle = {Medical Image Computing and Computer-Assisted Intervention -- MICCAI 2023},
  pages     = {178--188},
  publisher = {Springer},
  year      = {2023}
}

@article{wu2023meru,
  author  = {Wu, Fuping and Zhuang, Xiahai},
  title   = {Minimizing Estimated Risks on Unlabeled Data: A New Formulation for Semi-Supervised Medical Image Segmentation},
  journal = {IEEE Transactions on Pattern Analysis and Machine Intelligence},
  volume  = {45},
  number  = {5},
  pages   = {6021--6036},
  year    = {2023}
}

@inproceedings{bassi2025learning,
  author    = {Bassi, Pedro R. A. S. and Li, Wenxuan and Chen, Jieneng and Zhu, Zheren and Lin, Tianyu and Decherchi, Sergio and Cavalli, Andrea and Wang, Kang and Yang, Yang and Yuille, Alan L. and Zhou, Zongwei},
  title     = {Learning Segmentation from Radiology Reports},
  booktitle = {Medical Image Computing and Computer-Assisted Intervention -- MICCAI 2025},
  series    = {Lecture Notes in Computer Science},
  volume    = {15964},
  pages     = {305--315},
  publisher = {Springer},
  year      = {2025},
  doi       = {10.1007/978-3-032-04971-1_29}
}

@article{hu2025better,
  author  = {Hu, Liangqing and Meng, Zuqiang and Tan, Chaohong and Zhou, Yumin},
  title   = {Better Pseudo-Labeling for Semi-Supervised Domain Generalization in Medical Magnetic Resonance Image Segmentation},
  journal = {International Journal of Computational Intelligence Systems},
  year    = {2025},
  volume  = {18},
  number  = {1},
  pages   = {71},
  doi     = {10.1007/s44196-025-00786-8}
}

@article{gu2025sam,
  author  = {Gu, Hanxue and Dong, Haoyu and Yang, Jichen and Mazurowski, Maciej A.},
  title   = {How to Build the Best Medical Image Segmentation Algorithm Using Foundation Models: A Comprehensive Empirical Study with Segment Anything Model},
  journal = {Machine Learning for Biomedical Imaging},
  year    = {2025},
  volume  = {3},
  pages   = {88--120},
  doi     = {10.59275/j.melba.2025-86a6}
}

@article{cheng2025dynamic,
  author  = {Cheng, Zhiming and Liu, Mingxia and Yan, Chenggang and Wang, Shuai},
  title   = {Dynamic Domain Generalization for Medical Image Segmentation},
  journal = {Neural Networks},
  year    = {2025},
  volume  = {184},
  pages   = {107073},
  doi     = {10.1016/j.neunet.2024.107073}
}

@inproceedings{chen2025grata,
  author    = {Chen, Ziyang and Ye, Yiwen and Pan, Yongsheng and Xia, Yong},
  title     = {Gradient Alignment Improves Test-Time Adaptation for Medical Image Segmentation},
  booktitle = {Proceedings of the AAAI Conference on Artificial Intelligence},
  volume    = {39},
  pages     = {2429--2437},
  year      = {2025},
  doi       = {10.1609/aaai.v39i3.32244}
}

@inproceedings{ji2026tact,
  author    = {Ji, Jianhang and Cheng, Zhiming and Zhao, Jianxiang and Ma, Bingtao and Chen, Hao and Gao, Yuhan and Zhang, Lian and Ying, Zuobin and Wang, Shuai},
  title     = {Wake the Sleeping Weights: Sparsely-Activated Continual Test-Time Adaptation for Medical Image Segmentation},
  booktitle = {Proceedings of the IEEE/CVF Conference on Computer Vision and Pattern Recognition Findings},
  pages     = {7799--7809},
  year      = {2026}
}

@article{chen2025gaze,
  title     = {From gaze to insight: Bridging human visual attention and vision language model explanation for weakly-supervised medical image segmentation},
  author    = {Chen, Jingkun and Duan, Haoran and Zhang, Xiao and Gao, Boyan and Grau, Vicente and Han, Jungong},
  journal   = {IEEE Transactions on Medical Imaging},
  year      = {2025},
  publisher = {IEEE}
}

@article{chen2025addressing,
  title     = {Addressing inconsistent labeling with cross image matching for scribble-based medical image segmentation},
  author    = {Chen, Jingkun and Huang, Wenjian and Zhang, Jianguo and Debattista, Kurt and Han, Jungong},
  journal   = {IEEE Transactions on Image Processing},
  volume    = {34},
  pages     = {842--853},
  year      = {2025},
  publisher = {IEEE}
}

@inproceedings{lou2026cosseg,
  title     = {{CoSSeg-TTA}: Contrast-Aware Semi-Supervised Segmentation with Domain Generalization and Test-Time Adaptation},
  author    = {Lou, Jincan and Chen, Jingkun and Li, Haoquan and Li, Hang and Huang, Wenjian and Chen, Weihua and Wang, Fan and Zhang, Jianguo},
  booktitle = {MICCAI Challenge on Comprehensive Analysis and Computing of Real-World Medical Images},
  pages     = {57--67},
  year      = {2026},
  organization = {Springer}
}

\end{document}